\documentclass{article}
\usepackage{spconf,amsmath,graphicx,hyperref,array,booktabs,caption,multirow}
\usepackage{colortbl,xcolor,tipa}
\usepackage{tcolorbox}
\usepackage{colortbl}
\usepackage{cleveref}
\usepackage{amssymb}

\title{LoRALib: A Standardized Benchmark for Evaluating LoRA-MoE Methods}
\name{\parbox{\linewidth}{\centering Shaoheng Wang$^*$\textsuperscript{1,3}\thanks{* Equal contribution} \quad Yao Lu$^*$\textsuperscript{1,3,4} \quad Yuqi Li\textsuperscript{2}  \quad Yaxin Gao\textsuperscript{1,3} \quad Jiaqi Nie\textsuperscript{1,3} \\ \quad Shanqing Yu\textsuperscript{1,3} \quad Yingli Tian\textsuperscript{2} \quad Qi Xuan\textsuperscript{1,3}}}
\address{\textsuperscript{1}{Zhejiang University of Technology} \quad \textsuperscript{2}{The City University of New York}\\
\quad \textsuperscript{3}{Binjiang Institute of Artificial Intelligence, Zhejiang University of Technology} 
\quad \textsuperscript{4}{A*STAR}
}

\begin{document}
%
\maketitle
\begin{abstract}
As a parameter efficient fine-tuning (PEFT) method, low-rank adaptation (LoRA) can save significant costs in storage and computing, but its strong adaptability to a single task is often accompanied by insufficient cross-task generalization capabilities. To improve this, existing work combines LoRA with mixture-of-experts (MoE) to enhance the model's adaptability through expert modules and routing mechanisms. However, existing LoRA–MoE methods lack unified standards in models, datasets, hyperparameters, and evaluation methods, making it difficult to conduct fair comparisons between different methods. To this end, we proposed a unified benchmark named LoRALib. Specifically, we standardized datasets from $40$ downstream tasks into a unified format, fine-tuned them using the same hyperparameters and obtained $680$ LoRA modules across $17$ model architectures. Based on this LoRA library, we conduct large-scale experiments on $3$ representative LoRA-MoE methods and different LoRA selection mechanisms using the open-sourced testing tool OpenCompass. Extensive experiments show that LoRAMoE performs best, and that prioritizing LoRAs relevant to the target task can further improve the performance of MoE. We hope these findings will inspire future work. Our datasets and LoRA library are available at \url{https://huggingface.co/datasets/YaoLuzjut/LoRAOcean_dataset} and \url{https://huggingface.co/YaoLuzjut/models}.

\end{abstract}
\section{Introduction}
In recent years, large language models (LLM) have achieved remarkable success in many tasks such as logical reasoning~\cite{zhang2025rearank,liang2024self,wang2025see}, text generation~\cite{brown2020language,wang2025unitmge} and code generation~\cite{wang2024enhancing,huang2024bias}. However, the billions of model parameters make it impractical to fine-tune them completely. To this end, researchers have proposed a variety of PEFT methods to reduce the training cost. Among them, LoRA~\cite{hu2022lora} injects low-rank matrices into the original model layers, and can achieve excellent results close to full fine-tuning by training only a very small number of parameters, which has attracted widespread attention in the research community. Despite this, LoRA-based methods face challenges in handling multiple tasks simultaneously~\cite{xin2024beyond}, which hampers cross-task generalization in LLMs.

To mitigate this, some approaches attempt to combine LoRA with MoE by introducing several LoRA modules as experts and integrating them using a router network. For example, LoRAMoE~\cite{dou2024loramoe} is an MoE-style plugin and introduces localized balancing constraint during the training phase to alleviate the world knowledge forgetting. X-LoRA~\cite{buehler2024x} utilizes the hidden states to dynamically mix LoRA modules, enabling drawing upon different capabilities to solve tasks. Rocket~\cite{yang2024leveraging} builds a MoE system to make the best use of individual-yet-complementary knowledge between multiple LoRAs. Although existing methods have achieved promising results, they lack unified standards in terms of models, datasets, hyperparameters, and evaluation frameworks, making it difficult to make fair comparisons between different methods.

To this end, we propose a unified benchmark LoRALib to enable a fair comparison of various LoRA–MoE methods. First, we unify the datasets from $40$ downstream tasks into the same data format and train $680$ LoRA modules on $17$ different model architectures, providing sufficient LoRA samples for subsequent LoRA-MoE method comparison. Subsequently, we use the constructed LoRA library to conduct systematic experiments on $3$ typical LoRA–MoE frameworks, including LoRAMoE, X-LoRA and Rocket. To ensure the standardization and reproducibility of the evaluation, all experiments are performed through the open-sourced OpenCompass\footnote{\url{https://github.com/open-compass/opencompass}}. Extensive experiments show that LoRAMoE outperforms the other two methods. Besides, we further verify the impact of different LoRA selection strategies on the performance of LoRA–MoE and find that prioritizing LoRAs that are relevant to the target task consistently improves MoE performance.

Concretely, our contributions are as follows:
\begin{itemize}
    \item We build a large-scale open-sourced LoRA library based on a unified data format and training hyperparameters. This library contains $680$ LoRA modules fine-tuned on $17$ different model architectures on $40$ downstream tasks.
    \item Built upon this LoRA library, we systematically compare three types of LoRA–MoE implementations (LoRAMoE, X-LoRA and Rocket) on the open-sourced evaluation framework OpenCompass. Experimental results show that LoRAMoE achieves the best overall performance on most tasks and models.
    \item We further investigate the impact of different LoRA selection strategies on MoE performance. By comparing various strategies, including perplexity-based (PPL) selection, random sampling (Random), and manual selection (Manual), we find that prioritizing LoRAs highly relevant to the target task (i.e., Manual) consistently improves MoE's overall performance. This gain is particularly obvious for larger models and tasks requiring complex reasoning.
\end{itemize}

\section{Method}
In this paper, we build a LoRA library and establish a unified benchmarking pipeline to compare different LoRA–MoE methods and LoRA selection strategies. The following sections describe the implementation details of the LoRA library and benchmarking pipeline. 

\subsection{LoRA Library Construction}
\textbf{Dataset.} Before building the LoRA library, we first standardize the format of the datasets used in the experiment. Specifically, the data to construct our LoRA library are summarized as follows:  MedQA, COPA, MMLU, MATH, DROP, ARC, KOBEST, KorMedMCQA, LogiQA2.0, MathQA, MELA, QASPER, PIQA, SciQ, SIQA, SWAG, QNLI, WinoGrande, LogiQA, GSM8K, BLiMP, ANLI, CommonsenseQA, HellaSwag, OpenBookQA, MuTual, NQ-Open, bAbI, ETHICS/V-irtue, ETHICS/Utilitarianism, ETHICS/Justice, ETHICS/De-ontology, ETHICS/Commonsense, GLUE/Muli, GLUE/Mrpc, GLUE/Cola, GLUE/Qnli, GLUE/Qqp, GLUE/Rte, GLUE/St-sb, GLUE/Sst2. We choose these datasets because they provide comprehensive coverage of mainstream tasks and domains. All raw datasets are processed into the following format: \{"instructions": $\langle$\texttt{instruction}$\rangle$, "inputs": $\langle$\texttt{input (can be empty)}$\rangle$, "outputs": $\langle$\texttt{output}$\rangle$\}. We provide a data example in \Cref{fig:dataformat}.

\textbf{Model.} In this paper, we select $17$ representative backbones to cover different architectures and scales. The chosen models span parameter sizes from approximately $0.5$B to $9$B, including Phis~\cite{abdin2024phi3technicalreporthighly} (Phi-1.5, Phi-2, Phi-3-mini-4K-Instruct, Phi-3.5-mini-Instruct), Qwens (Qwen1.5-0.5B, Qwen1.5-7B, Qwen2.5-0.5B), Llamas~\cite{grattafiori2024llama} (Llama-3.2-1B-Instruct, Llama-3.2-3B-Instruct, Llama-3-8B, Llama-3.1-8B-Instruct), Yis~\cite{young2024yi} (Yi-1.5-6B, Yi-1.5-9B, Yi-6B, Yi-9B), Mistral-7B-v0.2 and TinyLlama-1.1B.

\textbf{LoRA Fine-tuning.} For each dataset, we freeze the original weights $W_0\in\mathbb{R}^{d\times k}$ and constrain the update to a low-rank decomposition: 
\begin{equation}
    W_0 + \Delta W_0 = W_0 + B A, 
\end{equation}
where \(B\in\mathbb{R}^{d\times r}\) and \(A\in\mathbb{R}^{r\times d}\) with the rank \(r \ll \min(d,k)\). During training, the pre-trained
weights \(W_0\) are frozen and only the matrices \(A\) and \(B\) remain trainable.
Follow~\cite{ma2023llm}, we set the hyperparameters for optimization: a batch size of $64$, gradient accumulation steps of $16$, an epoch of $2$ and an initial learning rate of $1$e$-5$. For all LoRAs, the rank is set to $16$.



\begin{figure}[t]
	\centering
    \begin{tcolorbox}[colback=gray!10!white, colframe=gray!50!black, title=Dataset Template\label{box:filter}, fontupper=\small]
     Instructions:  Weng earns $\$12$ an hour for babysitting. Yesterday, she just did $50$ minutes of babysitting. How much did she earn?\\
     \\
    Inputs: \{Could be empty\}\\
    Outputs: Weng earns $12/60 = \$\texttt{<<}12/60=0.2\texttt{>>}0.2$ per minute. Working $50$ minutes, she earned $0.2 \times 50 =\$\texttt{<<}0.2*50=10\texttt{>>} 10$. \quad \#\#\#\# 10.\\
    \end{tcolorbox}
	\caption{Data example.}
\label{fig:dataformat}
\end{figure}

\begin{table*}[!t]
  \centering
  \caption{Performance comparison of $3$ different LoRA–MoE methods using the same LoRA experts.}
  \resizebox{0.99\textwidth}{!}{
    \begin{tabular}{cc|ccccccccc|c}
    \toprule
    \multicolumn{1}{c}{Model} & \multicolumn{1}{c|}{Method} & \multicolumn{1}{c}{QNLI} & \multicolumn{1}{c}{Commonsense} & \multicolumn{1}{c}{Justice} & \multicolumn{1}{c}{LogiQA} & \multicolumn{1}{c}{DROP} & \multicolumn{1}{c}{MMLU} & \multicolumn{1}{c}{GSM8K} & \multicolumn{1}{c}{MedQA} & \multicolumn{1}{c|}{ARC} & \multicolumn{1}{c}{Avg} \\
    \midrule
    \multicolumn{1}{c}{\multirow{3}[2]{*}{Llama-3.2-1B-Instruct}} & Rocket & 97.63  & 38.92  & 47.50  & 26.67  & 25.78  & 32.85  & 58.67  & 32.10  & 48.43  & 45.39  \\
          & LoRAMoE & 90.83  & 43.67  & 47.50  & 30.87  & 28.93  & 37.65  & 64.57  & 37.40  & 50.29  & 47.97  \\
          & X-LoRA & 79.54  & 31.84  & 45.50  & 25.97  & 22.59  & 34.43  & 51.68  & 30.57  & 42.49  & 40.51  \\
    \midrule
    \multicolumn{1}{c}{\multirow{3}[2]{*}{Llama-3.2-3B-Instruct}} & Rocket & 99.32  & 55.11  & 47.50  & 30.87  & 27.98  & 38.23  & 67.01  & 37.07  & 60.65  & 51.53  \\
          & LoRAMoE & 98.45  & 52.82  & 49.50  & 34.90  & 30.74  & 38.96  & 70.43  & 41.59  & 64.45  & 53.54  \\
          & X-LoRA & 83.10  & 43.96  & 47.50  & 26.95  & 24.48  & 36.47  & 58.07  & 33.66  & 42.78  & 44.11  \\
    \midrule
    \multicolumn{1}{c}{\multirow{3}[2]{*}{Llama3.1-8B-Instruct}} & Rocket & 99.98  & 63.96  & 47.50  & 29.04  & 30.67  & 44.56  & 78.84  & 46.89  & 71.21  & 56.96  \\
          & LoRAMoE & 99.67  & 68.54  & 54.50  & 39.09  & 35.01  & 51.37  & 80.47  & 51.69  & 76.53  & 61.87  \\
          & X-LoRA & 84.90  & 60.06  & 50.50  & 31.50  & 31.59  & 41.58  & 64.80  & 40.72  & 56.71  & 51.37  \\
    \midrule
    \multicolumn{1}{c}{\multirow{3}[2]{*}{Qwen1.5-0.5B}} & Rocket & 84.38  & 29.82  & 45.50  & 25.41  & 24.46  & 29.17  & 53.68  & 29.42  & 40.12  & 40.22  \\
          & LoRAMoE & 83.70  & 39.20  & 42.50  & 29.31  & 26.20  & 34.50  & 59.50  & 34.10  & 47.20  & 44.02  \\
          & X-LoRA & 75.60  & 27.30  & 43.50  & 24.20  & 20.70  & 30.60  & 48.20  & 27.10  & 38.20  & 37.27  \\
    \midrule
    \multicolumn{1}{c}{\multirow{3}[2]{*}{Mistral-7B-v0.2}} & Rocket & 99.16  & 62.37  & 47.50  & 29.12  & 30.28  & 43.80  & 77.71  & 45.52  & 69.89  & 56.15  \\
          & LoRAMoE & 99.60  & 68.20  & 55.00  & 39.50  & 35.50  & 51.60  & 81.00  & 50.80  & 76.90  & 62.01  \\
          & X-LoRA & 80.27  & 61.19  & 52.50  & 31.60  & 31.70  & 42.00  & 67.50  & 42.20  & 57.80  & 51.86  \\
    \bottomrule
    \end{tabular}}
  \label{tab:compare}%
\end{table*}%

\begin{table*}[htbp]
  \centering
  \caption{LoRA adapters selected by different LoRA selection mechanisms when GSM8K is used as the target dataset.}
  \resizebox{0.99\textwidth}{!}{
    \begin{tabular}{cccccccc}
    \toprule
    \multirow{2}[3]{*}{Target Dataset} & \multicolumn{5}{c}{PPL}               & \multirow{2}[3]{*}{Random} & \multirow{2}[3]{*}{Munual} \\
\cmidrule{2-6}          & \multicolumn{1}{c}{Llama-3.2-1B-Instruct} & Llama-3.2-3B-Instruct & \multicolumn{1}{c}{Llama3-8B-Instruct\newline{}} & Qwen1.5-0.5B & \multicolumn{1}{c}{Mistral-7B-v0.2\newline{}} &       &  \\
\hline
    \multicolumn{1}{c}{\multirow{4}[1]{*}{GSM8K}} &  LogiQA2.0 & MedGA &  LogiQA2.0 & \multicolumn{1}{c}{OpenBookQA} & MedGA & BLiMP &  MATH \\
          & MedGA &  LogiQA2.0 &  PIQA &  LogiQA2.0 & ETHICS/Utilitarianism & NQ-Open &  MathQ \\
          & LogiQA &  PIQA & MedGA & SciQ  &  MATH &  MATH & MMLU \\
          &  PIQA & OpenBookQA &  MATH & ETHICS/Virtue &  PIQA &  LogiQA2.0 & DROP \\
    \bottomrule
    \end{tabular}}
  \label{tab:lora modules}%
\end{table*}%

\begin{table*}[t]
  \centering
  \caption{Performance comparison of different LoRA selection strategies. The experiments are performed using the LoRAMoE method.}
  \resizebox{0.99\textwidth}{!}{
    \begin{tabular}{cc|ccccccccc|c}
    \toprule
    Model & \multicolumn{1}{c|}{Selection Mechanism} & \multicolumn{1}{c}{QNLI} & \multicolumn{1}{c}{Commonsense} & \multicolumn{1}{c}{Justice} & \multicolumn{1}{c}{LogiQA} & \multicolumn{1}{c}{DROP} & \multicolumn{1}{c}{MMLU} & \multicolumn{1}{c}{GSM8K} & \multicolumn{1}{c}{MedQA} & \multicolumn{1}{c|}{ARC} & \multicolumn{1}{c}{Avg} \\
    \midrule
    \multicolumn{1}{c}{\multirow{3}[2]{*}{Llama-3.2-1B-Instruct}} & PPL   & 89.73  & 42.87  & 45.00  & 30.64  & 27.98  & 36.24  & 63.57  & 36.76  & 51.34  & 47.13  \\
          & Manual & 90.83  & 43.67  & 47.50  & 30.87  & 28.93  & 37.65  & 64.57  & 37.40  & 50.29  & 47.97  \\
          & Random & 87.06  & 41.95  & 54.50  & 29.01  & 26.06  & 35.01  & 62.89  & 36.07  & 49.58  & 46.90  \\
    \midrule
    \multicolumn{1}{c}{\multirow{3}[2]{*}{Llama-3.2-3B-Instruct}} & PPL   & 98.32  & 50.67  & 49.50  & 32.85  & 31.04  & 38.60  & 69.41  & 40.68  & 63.37  & 52.72  \\
          & Manual & 98.45  & 52.82  & 49.50  & 34.90  & 30.74  & 38.96  & 70.43  & 41.59  & 64.45  & 53.54  \\
          & Random & 98.10  & 50.00  & 54.50  & 32.06  & 30.01  & 37.45  & 68.90  & 39.64  & 62.16  & 52.54  \\
    \midrule
    \multicolumn{1}{c}{\multirow{3}[2]{*}{Llama3.1-8B-Instruct}} & PPL   & 99.45  & 67.65  & 52.50  & 38.64  & 34.59  & 49.06  & 79.68  & 49.50  & 74.34  & 60.60  \\
          & Manual & 99.67  & 68.54  & 54.50  & 39.09  & 35.01  & 51.37  & 80.47  & 51.69  & 76.53  & 61.87  \\
          & Random & 99.04  & 65.39  & 52.50  & 38.00  & 33.50  & 48.07  & 78.50  & 48.95  & 73.30  & 59.69  \\
    \midrule
    \multicolumn{1}{c}{\multirow{3}[2]{*}{Qwen1.5-0.5B}} & PPL   & 82.50  & 38.51  & 40.50  & 28.50  & 25.50  & 33.50  & 58.31  & 33.50  & 46.48  & 43.03  \\
          & Manual & 83.70  & 39.20  & 42.50  & 29.31  & 26.20  & 34.50  & 59.50  & 34.10  & 47.20  & 44.02  \\
          & Random & 81.10  & 37.60  & 40.50  & 27.80  & 25.00  & 32.80  & 57.26  & 32.50  & 48.45  & 42.56  \\
    \midrule
    \multicolumn{1}{c}{\multirow{3}[2]{*}{Mistral-7B-v0.2}} & PPL   & 99.20  & 66.80  & 53.00  & 38.70  & 34.70  & 49.80  & 80.20  & 49.00  & 74.50  & 60.66  \\
          & Manual & 99.60  & 68.20  & 55.00  & 39.50  & 35.50  & 51.60  & 81.00  & 50.80  & 76.90  & 62.01  \\
          & Random & 98.90  & 65.10  & 53.00  & 38.20  & 33.90  & 48.90  & 79.30  & 48.60  & 73.80  & 59.97  \\
    \bottomrule
    \end{tabular}}
  \label{tab:selection}%
\end{table*}%

\subsection{Benchmarking LoRA–MoE Methods}
\label{sec:Benchmarking}
In this paper, we select $3$ representative LoRA–MoE methods as evaluation objects. Before the experiment, we briefly introduce each method as follows:

\textbf{LoRAMoE.} LoRAMoE uses multiple LoRA adapters as experts and introduces a router in the FFN layer of transformer to dynamically weight the experts: 
\begin{equation}
    o = W_0 x + \sum_{i=1}^N G(x)_i\, B_i A_i x, 
\end{equation}
where \(o\) denotes the output of the linear layer and \(x\) denotes its input. \(W_0\) is the pre-trained frozen parameters of the linear layer. \(G(x)\) is the gating function, \(N\) is the number of experts, A and B are two low-rank matrices of LoRA. Besides, authors also propose localized balancing constraint approach to prevent routing degradation.

\textbf{X-LoRA.} Similar to LoRAMoE, X-LoRA takes a set of pre-trained LoRA adapters as experts and creates a dynamic gating approach to extend individual LoRA adapters with individual token and layer granularity to facilitate mixing deep inside the model.

\textbf{Rocket.} Rocket first selects high-potential expert candidates from the LoRA library based on K-shot, then retrieves open-source instructions with similar and diverse semantics to K-shot for data augmentation. Finally, it assembles the candidate experts into a MoE and jointly fine-tunes the router and experts to obtain the final task expert.

For a fair comparison, we use the same $4$ LoRAs as experts for the following experiments. Specifically, given a target dataset and a pre-trained model, we manually select $4$ LoRAs with the same or similar domains from the LoRAs library as experts based on task relevance. Next, we feed these experts to $3$ types of LoRA–MoE methods and perform evaluation through the open-sourced OpenCompass. We evaluate performance on QNLI, ETHICS/Commonsense, ETHICS/Justice, LogiQA, DROP, MMLU, GSM8K, MedQA, and ARC, which are chosen because they together cover a diverse range of task types. Due to page limitations, we only present the results on Llama-3.2-1B-Instruct, Llama-3.2-3B-Instruct, Llama3.1-8B-Instruct, Qwen1.5-0.5B and Mistral-7B-v0.2 in the main text. As shown in \Cref{tab:compare}, LoRAMoE achieves the highest average precision on all given models. Specifically, LoRAMoE outperforms Rocket by approximately $2.0\%$ to $5.9\%$ on average precision. In contrast, the training-free X-LoRA method significantly lags behind the other two methods on all backbone models. \textit{In summary, we recommend LoRAMoE as the preferred method to improve the multi-task generalization of the LoRA module.}

\subsection{LoRA Selection Mechanism}
In \Cref{sec:Benchmarking}, we have studied which LoRA-MoE method works best. In a bank of LoRA modules, it is essential to appropriately select the most relevant task expert candidates. Therefore, we further study the impact of the LoRA selection strategy on the final MoE performance. Given the good performance of LoRAMoE, we use it as a unified MoE framework and select $4$ experts from the LoRA library using $3$ strategies: PPL, Random, and Manual. The specific definitions and operations of the $3$ strategies are as follows.


\textbf{PPL.} For the target dataset, we first merge each LoRA in the LoRA library into the given model using \Cref{eq:merge}. 
\begin{equation}
    W_{\text{merged}} = W_0 + B A.
    \label{eq:merge}
\end{equation}
Then we calculate the PPL of the model after LoRA merging using \Cref{eq:ppl}.
\begin{equation}
\mathrm{PPL}(W_{\text{merged}}) := \exp\Bigg(-\sum_{i=1}^{|y|} \log P\big(y_{i}\mid x,y_{<i}; W_{\text{merged}}\big)\Bigg),
\label{eq:ppl}
\end{equation}
where \(P\big(y_{i} \mid x, y_{<i}; W_{\text{merged}}\big)\) denotes the predicted probability of the \(i\)-th token \(y_{i}\) of \(y\) given the input sequence \(x\) and the preceding tokens \(y_{<i}\) under the fine-tuned model with the merged LoRA \(W_{\text{merged}}\), and \(|y|\) is the length of sequence \(y\). A lower PPL indicates better language-modeling quality. Therefore, we sort the candidate LoRAs from low to high according to PPL, and select the top $4$ with the lowest PPL as the experts for the target task. Many existing LoRA–MoE studies~\cite{do2023hyperrouter,wang2024auxiliary,do2025simsmoe} also select LoRA based on it.

\textbf{Random.} Randomly select $4$ LoRAs from the LoRA library according to uniform distribution.

\textbf{Manual.} We manually selects $4$ LoRAs that are most relevant to the target task based on the task description and domain knowledge. This simulates the selection strategy commonly used by professional engineers in real-world scenarios.

The experiment uses LoRAMoE as a unified MoE framework. The expert candidate pool comes from $680$ LoRA instances previously trained and saved on $17$ models and $40$ tasks. For each target dataset and selected model, we first select $4$ experts according to each strategy and then utilize them to train the MoE model. Besides, all hyperparameter settings are consistent with \Cref{tab:compare}. Finally, to ensure fairness, we employ OpenCompass to evaluate the performance of three LoRA–MoE approaches. Similar to \Cref{tab:compare}, we also evaluate performance on QNLI, Commonsense, Justice, LogiQA, DROP, MMLU, GSM8K, MedQA and ARC. As shown in \Cref{tab:selection}, $3$ LoRA selection strategies exhibit consistent overall performance across multiple models. Specifically, manual selection achieves the highest average score across all listed models. Manual selection results in an average precision improvement of approximately $0.8\%–1.4\%$ compared to PPL selection method, and an even more significant improvement (approximately $1.0\%–2.2\%$) compared to random sampling. 

\textbf{Result Analysis.} \Cref{tab:lora modules} reports the LoRA adapters selected by each selection mechanism when GSM8K is used as the target dataset. Different LoRA selection strategies produce significantly different sets of experts for GSM8K. GSM8K is a typical mathematical reasoning dataset, and the experts selected by the Manual strategy (MATH, MathQ, MMLU, and DROP) are almost all from mathematics or complex reasoning-related fields. This directly aligns with the GSM8K task requirements, providing the model with stronger inductive bias and task adaptability. Therefore, the manual strategy consistently achieves the best performance in experiments. In contrast, while PPL can select high-performing LoRAs, its selection results tend to favor general logical reasoning tasks such as LogiQA and PIQA and medical knowledge (MedGA), rather than strictly aligned mathematical reasoning. This explains why, while its performance is close to Manual, it consistently lags behind. Random selection may choose LoRAs that are unrelated to GSM8K, such as BLiMP (Benchmark of Linguistic Minimal Pairs) and NQ-Open (Open Domain Question Answering), resulting in significantly lower performance in most cases. In summary, \textit{we recommend using manual selection strategies on critical or high-value tasks to achieve optimal results.}

\vspace{-4mm}

\section{Conclusion}
In this paper, we present LoRALib, a unified benchmark for evaluating LoRA–MoE methods across diverse models and tasks. Specifically, we first build a large-scale open-sourced LoRA library based on the unified data format and training hyperparameters. This library contains $680$ LoRA modules fine-tuned on $17$ different model architectures on $40$ downstream tasks. Then built upon this LoRA library, we benchmark three types of LoRA–MoE implementations on the open-sourced evaluation framework OpenCompass and find that LoRAMoE achieves the best overall performance. We further investigate the impact of different LoRA selection strategies on MoE performance and conclude that prioritizing LoRAs that are highly relevant to the target task (i.e., manual) can consistently improve the overall performance of MoE. We hope that our benchmark and findings will serve as a foundation and inspiration for future research on effective LoRA–MoE methods.

\clearpage

\bibliographystyle{IEEEbib}
\bibliography{strings,refs}

\clearpage
\appendix

\end{document}